# Improving Document-Level Sentiment Classification Using Importance of Sentences

**Gihyeon Choi [1], Shinhyeok Oh [1] and Harksoo Kim [2,\***

[1] Program of Computer and Communications Engineering, College of IT, Kangwon National University, Chuncheon-si 24341, Korea; pluto32@kangwon.ac.kr (G.C.); osh7605@kangwon.ac.kr (S.O.)
[2] Division of Computer Science and Engineering & Department of Artificial Intelligence, College of Engineering, Konkuk University, 120 Neungdong-ro, Gwangjin-gu, Seoul 05029, Korea
\* Correspondence: nlpdrkim@konkuk.ac.kr; Tel.: +82-2-450-3499

**Abstract:** Previous researchers have considered sentiment analysis as a document classification task, in which input documents are classified into predefined sentiment classes. Although there are sentences in a document that support important evidences for sentiment analysis and sentences that do not, they have treated the document as a bag of sentences. In other words, they have not considered the importance of each sentence in the document. To effectively determine polarity of a document, each sentence in the document should be dealt with different degrees of importance. To address this problem, we propose a document-level sentence classification model based on deep neural networks, in which the importance degrees of sentences in documents are automatically determined through gate mechanisms. To verify our new sentiment analysis model, we conducted experiments using the sentiment datasets in the four different domains such as movie reviews, hotel reviews, restaurant reviews, and music reviews. In the experiments, the proposed model outperformed previous state-of-the-art models that do not consider importance differences of sentences in a document. The experimental results show that the importance of sentences should be considered in a document-level sentiment classification task.

## 1. Introduction

Sentiment analysis is a natural language processing (NLP) task in which a given text is classified into predefined classes (e.g., positive, neutral, and negative). The initial models on sentiment analysis use hand-made sentiment lexicons that contain sentiment words annotated with polarities [1–3]. In general, they extract sentiment words from sentences. Based on discrete information, such as polarities and strengths of sentiment words, they classify sentences into sentiment classes with the strongest polarities [2–4]. Although these lexicon-based models are simple and efficient, they suffer from limitations. First, the manual construction of sentiment lexicons is a time-consuming and labor-intensive job. To overcome these limitations, some models to automatically construct sentiment lexicons have been proposed [5,6]. Second, a fixed polarity with strength should be assigned to each sentiment word although it may have different polarities depending on application domains. For example, "The air conditioner is so hot" expresses a negative opinion because "hot" here means "having mechanical trouble." By contrast, "The movie is so hot" expresses a positive opinion because "hot" here means "popular." To overcome these limitations, some models based on machine learning (ML) have been proposed [7]; however, these ML-based models need a large amount of data annotated with polarities for training. With the tremendous growth of user-generated corpus in rating scores, such as movie and goods reviews, various models based on deep neural networks (DNNs) have been proposed [8–10]. Although these DNN-based models show good performances, most of them do not consider the importance of each sentence in an input text; instead, they treat the input text as a bag of sentences. However, while a human determines polarity of a document, he/she reads through the whole document, removes ordinary sentences (i.e., unimpressive sentences), and determines the final polarity based on some impressive sentences. Table 1 summarizes an example of sentiment analysis in a movie review domain.

**Table 1.** Example of a movie review

| Sentence | Polarity of a sentence | Polarity of a document |
|---|---|---|
| "From my opinion, *No Country for Old Men* isn't the best weak Coen brothers film." | Weak negative | Strong positive |
| "Josh Brolin is hunting in the desert." | Neutral | |
| "But, that is not to say that it's a bad film." | Positive | |
| "It really is a solid piece of cinema." | Strong positive | |

As summarized in Table 1, although a document has only one polarity, each sentence in the document has different polarities. The first sentence has a polarity opposite to that of the document, and the second sentence does not have a biased polarity. The third and fourth sentences weakly and strongly affected the polarity of the document, respectively. In other words, the third and fourth sentences are strong evidences that support the polarity of the whole document. Therefore, to effectively determine polarity of a document, each sentence in the document should be dealt with different degrees of importance. To tackle this problem, we propose a DNN-based document-level sentiment classification model to automatically reflect the sentence importance meaning on how much each sentence supports polarity of a whole document. Then, we verify that considering sentence importance contributes to improve performances of document-level sentiment classification through experimental comparisons.

The remainder of this paper is organized as follows. In Section 2, we describe previous studies on sentiment analysis. In Section 3, we present a DNN model for document-level sentiment classification. In Section 4, we elaborate on the experimental setup and results. In Section 5, we discuss our experimental results. Finally, in Section 6, we conclude the study.

**2. Previous Work**

Previous studies on sentiment analysis are divided into lexicon-based and ML-based models. The lexicon-based models first define sentiment lexicons that contain sentiment words and their attributes, such as polarities and their strengths [2,3]. Then, they predict the sentiment classes of given texts using the number of sentiment words, total strength of the sentiment words, and maximum strength of the sentiment words [2–4]. Taboada et al. [11] proposed a rule set for calculating the polarities of input texts based on sentiment words and their surrounding linguistic clues, such as amplifiers (e.g., very, extraordinary, and most), downtoners (e.g., slightly, somewhat, and pretty), and negators (e.g., nobody, none, and nothing). The lexicon-based models have several advantages, such as being simple, deterministic, and efficient. However, they suffer from the fixed polarity problem as each word in a sentiment lexicon has a fixed polarity, although the polarities of words can be changed depending on domains. To tackle this issue, ML-based models have been proposed. These models predict the sentiment classes of given texts based on statistical or algebraic information obtained from a large amount of training corpus. Kim [8] proposed a sentiment classification model based on convolutional neural networks (CNNs) in which pre-trained word vectors for sentence classification tasks are used as inputs. Ren et al. [12] proposed a CNN-based model for the sentiment classification of texts in a social network service (SNS). To improve performances in an SNS domain, they used contextualized features (e.g., conversation-based, author-based, and topic-based contexts about a target text) that are well designed for SNS texts. To well reflect the information of word sequences in texts, some researchers [13–15] adopted recurrent neural networks (RNNs) that are effective in capturing long dependencies between words. Teng et al. [13] proposed an RNN model to automatically learn the rule set (i.e., sentiment strength, intensification, and negation of lexicon sentiments) proposed by Taboada et al. [11]. Recently, BERT (a large-scale pre-trained language model) showed the best performances in various downstream NLP tasks through fine-tuning [16]. Hoang et al. [17] showed that using the contextual word representations from BERT is effective in a sentiment analysis task. To capture sentiment signals over hierarchical phrase structures, sentence representation methods based on tree-structured RNNs were proposed [18,19]. Many previous ML-based models are focused on sentence-level sentiment classification. In document-level sentiment classification (i.e., sentiment classification on a document containing multiple sentences), they consider the text as a bag of sentences without considering the importance of each sentence. Thongtan and Phienthrakul [20] proposed a neural embedding model to obtain document embeddings using cosine similarity instead of the dot product. Abdi et al. [21] investigated a method to effectively merge multi-features such as word embedding, sentiment knowledge, sentiment shifter rules, and linguistic knowledge in order to overcome flaws raised by flat concatenation of different features. However, these approaches on document-level sentiment analysis cannot consider how much each sentence contributes in determining the polarity of a given document. To overcome this limitation, we propose a DNN-based sentiment classification model in which sentences in a document differently contribute to document-level classification according to their importance.

## 3. Document-Level Sentiment Analysis Model

As shown in Figure 1, the proposed model consists of three submodules: A sentence encoder, a document encoder, and a sentiment classifier.

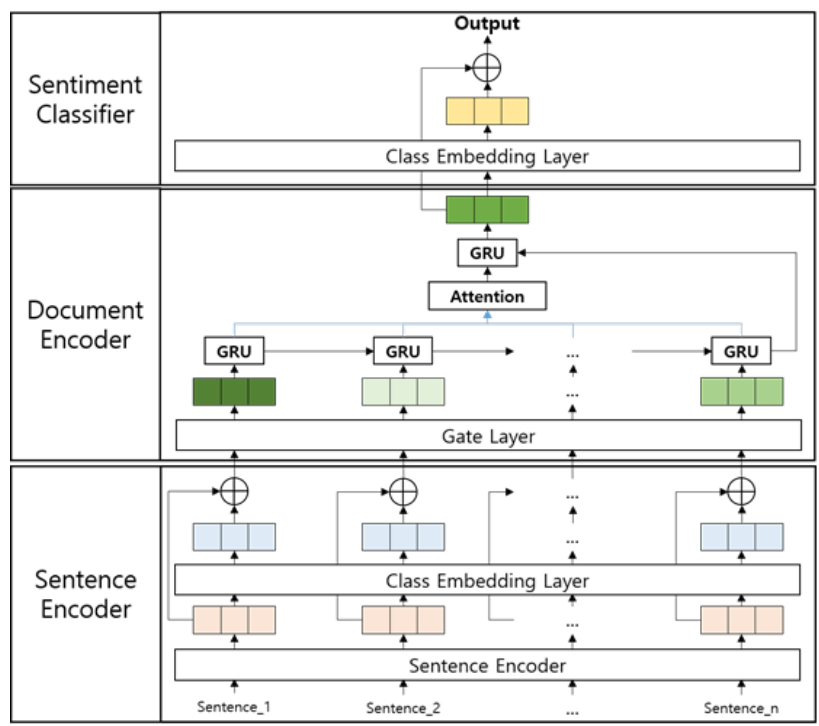

**Figure 1.** Overall architecture of the proposed model

The sentence encoder generates the embeddings of each sentence in a given document using ALBERT (a light version of BERT) [22]. Then, it enriches the sentence embeddings by adding the embeddings of sentiment classes. The document encoder calculates the importance of each sentence through gate functions. Then, it generates a document embedding by weighted summing the sentences according to the calculated importance. Subsequently, it enriches the document embedding by adding the embeddings of sentiment classes. The enriched document embedding is used as an input of the sentiment classifier. The sentiment classifier determines a sentiment class of the input document through a fully connected neural network (FNN).

*3.1. Sentence Encoder*

The sentence encoder converts each input sentence into embedding vectors. To obtain sentence embeddings, including contextual information, we adopt ALBERT, as shown in Figure 2.

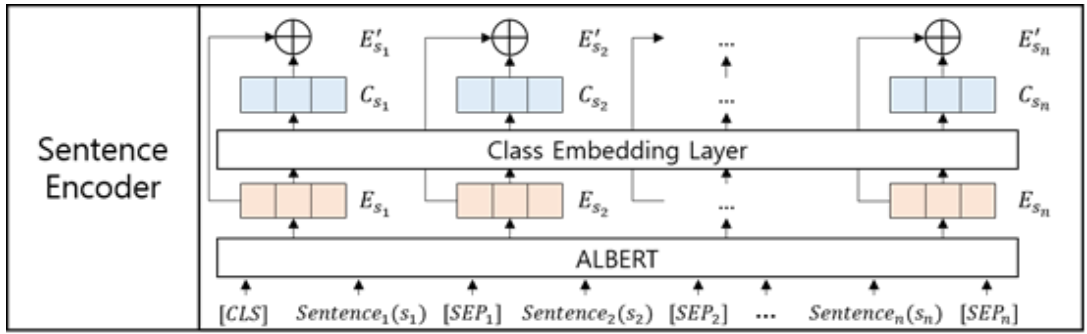

**Figure 2.** Architecture of the sentence encoder

In Figure 2, $s_i$ is the *i*-th sentence in which each word is represented as ALBERT embeddings. Then, *[CLS]* and *[$SEP_i$]* are the class token (i.e., a special token for classification tasks) and *i*-th separator token (i.e., a special token for representing a sentence boundary between the *i*-th sentence and *i+1*-th sentence), respectively. In language models, such as BERT and ALBERT, the output vector of the class token generally conveys the task-oriented meaning of an input document (i.e., a sequence of all input words). Accordingly, we assume that the output vectors of the separator tokens convey the representations of each input sentence, as reported by Cohan et al. [23]. Therefore, the input sentences $s_1, s_2, ..., s_n$ in a document are converted into the sentence embeddings $E_{s_1}, E_{s_2}, ..., E_{s_n}$, as shown in the following equation:

$$E_{s_1}, E_{s_2}, ..., E_{s_n} = ALBERT(s_1, s_2, ..., s_n), \qquad (1)$$

where $E_{s_i}$ denotes an output vector of the *i*-th separator token. To supplement the sentence embeddings with the domain knowledge of sentiment classes, we adopt a domain embedding scheme proposed in [24], as shown in the following equation:

$$C_{s_i} = W_c \cdot FNN(E_{s_i}), \qquad (2)$$

where $FNN(E_{s_i})$ denotes an FNN with rectified linear activation unit (ReLU) [25] output functions for mapping the *i*-th sentence embedding $E_{s_i}$ into the vector space of target classes (i.e., a positive class and a negative class in a sentiment classification task), and $W_c$ denotes a weight matrix that consists of randomly initialized embeddings of target classes. Then, $C_{s_i}$ denotes a class similarity embedding containing inner product values between the *i*-th input sentence and each target class. The class similarity embedding represents the degrees of association between an input sentence and target classes. Finally, the sentence encoder generates a domain-specific sentence embedding $E'_{s_i}$ of the *i*-th sentence by concatenating the general sentence embedding $E_{s_i}$ and the class similarity vector $C_{s_i}$, as shown in the following equation:

$$E'_{s_i} = [E_{s_i}; C_{s_i}]. \qquad (3)$$

*3.2. Document Encoder*

The document encoder calculates the importance degree of each sentence in resolving a given sentiment classification problem. Then, it generates a document embedding based on the importance degrees of each sentence through gated recurrent units (GRUs) [26] and an attention mechanism [27], as shown in Figure 3.

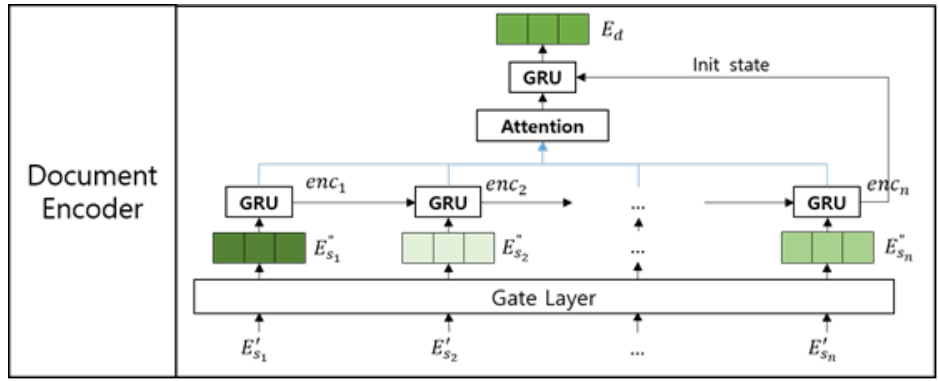

**Figure 3.** Architecture of the document encoder

To calculate the importance degrees of each sentence, we adopt a gate mechanism, as shown in the following equation:

$$g_i = \sigma(W_g \cdot E'_{s_i})$$
$$E''_{s_i} = g_i \cdot E'_{s_i}, \qquad (4)$$

where $W_g$ denotes a randomly initialized weight matrix, $\sigma$ denotes a sigmoid function for calculating the importance degree of $E'_{s_i}$, and $E''_{s_i}$ is a sentence embedding in which the importance degree is reflected. Then, the document encoder encodes all the gated sentence embeddings using a GRU encoder, as expressed by the following equation:

$$enc_i = GRU(E''_{s_i}, enc_{i-1}), \qquad (5)$$

where $enc_i$ denotes the *i*-th gated sentence embedding encoded by the forward state of the GRU. Then, the document encoder generates a document embedding using a GRU decoder with Luong's attention mechanism [24], as shown in Figure 4.

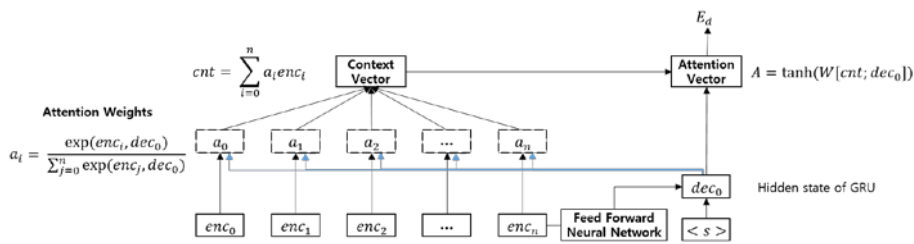

**Figure 4.** Gated recurrent unit (GRU) decoder with Luong's attention mechanism.

As shown in Figure 4, each attention weight $a_i$ is induced by inner products between each output $enc_i$ of the GRU encoder and the first hidden state $dec_0$ of the RNN decoder. The attention weights signify how much each output $enc_i$ of the GRU encoder is associated with the first hidden state $dec_0$ of the GRU decoder. Then, the context vector $cnt$ is constructed by the weighted sum of $a_i$ and $enc_i$. Finally, the RNN decoder generates a document embedding $E_d$ using the last hidden state $enc_n$ of the RNN encoder, start symbol $<s>$, and context vector $cnt$, as expressed by the following equation:

$$E_d = GRU(FNN(enc_n), <s>, cnt). \tag{6}$$

*3.3. Sentiment Classifier*

To determine the sentiment class of an input document, the sentiment classifier uses the document embedding and a class similarity embedding as inputs. The class similarity embedding is the same as that in the sentence encoder except that the class similarity embedding represents the degrees of association between the document embedding and target classes, as expressed by the following equation:

$$C_d = W_c \cdot FNN(E_d), \tag{7}$$

where $FNN(E_d)$ denotes an FNN with ReLU output functions for mapping the document embedding $E_d$ into the vector space of target classes, and $W_c$ is the same class embedding matrix as that in Equation (2). In other words, the sentiment classifier shares the class embedding matrix with the sentence encoder. Then, $C_d$ denotes a class similarity embedding containing inner product values between the document embedding and each target class. Finally, the sentiment classifier determines the sentiment class of an input document through an FNN with sigmoid output functions, as expressed by the following equation:

$$Sentiment(s_1, s_2, ..., s_n) = FNN([E_d; C_d]), \tag{8}$$

where $[E_d; C_d]$ denotes a concatenation of the document embedding $E_d$ and class similarity embedding $C_d$.

## 4. Results

*4.1. Datasets and Experimental Settings*

For our experiments, we used the IMDB dataset (135,669 documents) [28], the Yelp-hotel dataset (34,961 documents) [29], the Yelp-rest dataset (178,239 documents) [29], and the Amazon dataset (83,159 documents) [29]. The IMDB dataset is a movie review dataset annotated with 10-scale polarities. Then, the Yelp-hotel dataset, the Yelp-rest dataset, and the Amazon dataset are a hotel review dataset, a restaurant dataset, and a music review dataset that are annotated with 5-scale polarities, respectively. Table 2 lists data statistics of the four dataset.

**Table 2.** Data statistics of experiment datasets. We show the number of documents in each split.

| Dataset | Train | Development | Test |
|---|---|---|---|
| IMDB | 108,535 | 13,567 | 13,567 |
| Yelp-hotel | 20,975 | 6,993 | 6,993 |
| Yelp-rest | 106,943 | 35,648 | 35,648 |
| Amazon | 59,399 | 11,880 | 11,880 |

For fair comparison with the previous models, we encoded review scores of the Yelp-hotel dataset, the Yelp-rest dataset, and the Amazon dataset into three discrete categories (score >3 as positive, =3 as neutral, and <3 as negative) according to Huang and Paul's experimental settings [29].

For sentence segmentation, we used NLTK [30], an open source Python library for NLP. Then, we evaluated the proposed model based on accuracy, as expressed by the following equation.

$$Accuracy = \frac{\text{\# of correct polarities}}{\text{\# of polarities returned by a system}}. \tag{9}$$

We implemented the proposed model using PyTorch [31]. The training and prediction were performed on a per-document level. Table 3 lists the parameter settings for the model training.

Table 3. Optimal hyperparameters

| Parameter | Value |
|---|---|
| Dimension of token embedding | 128 |
| Dimension of a hidden node in the class embedding | 300 |
| Dimension of q hidden nodes in the sentence encoder | 768 |
| Max sentence length | 512 |
| Max number of sentences | 50 |
| Batch size | 64 |
| Learning rate | 0.00002 |

*4.2. Experiments*

The first experiment compared the performances of the proposed model with those of the previous models. Table 4 lists the comparison results. In Table 4, Kim-CNN [8] is a sentence classification model that shows good performances, although it uses simple CNNs. Adhikari-logistic regression [28] and Adhikari-support vector machine [28] are text classification models based on logistic regression and support vector machine, in which the term frequency and inversed document-frequency scores are used as features, respectively. HAN [32] extracts meaningful features by modeling the hierarchical structure of a document and classifies the document into predefined classes using two levels of attention mechanisms: Word-level attentions and sentence-level attentions. LSTM-Reg [28] is a sentiment classification model based on single-layer bidirectional long short-term memory (BiLSTM). Knowledge distillation (KD)-LSTM [28] is a modified version of LSTM that uses the KD scheme to increase performances using fine-tuned BERT-Large [16]. ALBERT-Base is our baseline model, in which sentiment classification is performed using only ALBERT [22]. ALBERT showed state-of-the-art performances in many downstream NLP tasks, such as span prediction, sequence labeling, and text classification.

Table 4. Performance comparison with the previous models

| Model | IMDB | | Yelp-hotel | | Yelp-rest | | Amazon | |
|---|---|---|---|---|---|---|---|---|
| | Valid | Test | Valid | Test | Valid | Test | Valid | Test |
| Kim-CNN [8] | 0.429 | 0.427 | 0.794 | 0.775 | 0.805 | 0.806 | 0.853 | 0.817 |
| Adhikari-support vector machine [28] | 0.425 | 0.424 | - | | - | | - | |
| Adhikari-logistic regression [28] | 0.431 | 0.434 | - | | - | | - | |
| HAN [32] | 0.518 | 0.512 | 0.833 | 0.810 | 0.841 | 0.839 | 0.867 | 0.848 |
| ALBERT-Base | 0.520 | 0.519 | 0.827 | 0.827 | 0.871 | 0.874 | 0.870 | 0.858 |
| LSTM-Reg [28] | 0.534 | 0.528 | 0.813 | 0.796 | 0.837 | 0.840 | 0.863 | 0.837 |
| Knowledge distillation-LSTM [28] | 0.545 | 0.537 | - | | - | | - | |
| Proposed model | 0.546 | 0.548 | 0.843 | 0.833 | 0.878 | 0.882 | 0.885 | 0.876 |

As summarized in Table 4, the proposed model outperformed all the previous sentiment classification models in the experiments with all the datasets. To statistically validate the performance differences, we performed t-tests between the proposed model and the comparison models using the accuracies as the input values of the t-test. The p-values were 0.000311 between the proposed model and Kim-CNN [8], 0.000164 between the proposed model and HAN [32], 0.000164 between the proposed model and ALBERT-Base, and 0.001341 between the proposed model and LSTM-Reg [28], respectively. This implies that the performances are statistically meaningful at a significance level of 99%. These experimental results show that a well-formed neural network architecture has better performances, effectively reflecting the importance of sentence and class information for document-level sentiment classification.

Second, we evaluated the effectiveness of each module in the proposed model at the architecture level by using the IMDB dataset. The experimental results are summarized in Table 5.

**Table 5.** Performance comparison depending on changes in the architecture

| Model | Accuracy (Accuracy in the valid. dataset) |
| --- | --- |
| The whole model | 0.548 (0.546) |
| - Class similarity embedding for a sentence | 0.545 (0.545) |
| - Gated sentence embedding | 0.543 (0.543) |
| - Class similarity embedding for a document | 0.545 (0.548) |

In Table5, "The whole model" is our model, as shown in Figure 1. "Class similarity embedding for a sentence" is a modified version of our model, in which is equal to, by excluding Equation (2), and "Gate sentence embedding" is a modified version of our model, in which is equal to, by excluding Equation (4). Then, "Class similarity embedding for a document" is a modified version of our model in which Equation (7) is excluded. As summarized in Table 5, the modified versions showed inferior performances, compared with the whole model. This fact reveals that the proposed embedding methods (i.e., class similarity embedding and gated sentence embedding) contribute to improve the performances of the document sentence classification. Moreover, "Gate sentence embedding" showed the biggest performance drop in both the test dataset and the validation dataset. This fact reveals that the importance of sentences should be considered in a document-level sentiment classification task.

## 5. Discussion

To check whether the importance of sentence is learned through the proposed neural network architecture or not, we visualized the importance degrees (i.e., scores of the gate function in Equation (4)) of each sentence in the test documents (i.e., documents in the test dataset) through two-dimensional heat maps, as shown in Figure 5.

| Example of positive review | Example of negative review |
| --- | --- |
| the force is back big time, a slow start may cause tripidation, but after twenty minutes we are off, the first chase with anakin and obi wan after an assassin sets the pace, which never really lets up. i am not going to go into plot details as this would ruin it for anyone who has not seen it. but just to say there are scenes which make you leap out of your seat. a great tag line would be more light sabres than you could shake a stick at go and see it i shall be going at least another three times, as for the usual boring critics, who never understood the films ignore them | Sure, we can forgive the lack of realism in the stunts and the cheesiness of the action scenes. But the movie seemed like a satire of the old Indy movies. There was no chemistry between the actors, the action sequences were long and uninteresting, and put together it was not a very well made film. Moviegoers shouldn't be told that our expectations are off to compensate for a low-rate movie, we should be commended for warning others of this debacle. This movie milked millions of dollars primarily because of reputation and not of substance, and that's what is upsetting. For the many who support this movie for whatever reason, realize that both your time and your money is on the line, neither of which should be compromised for nostalgia. |

**Figure 5.** Heat map for visualizing sentence importance

In Figure 5, the sentences more associated with polarity of the given document were colored in bluer. As shown in Figure 5, each sentence differently contributes to determine document-level polarity. Then, we computed standard deviations of min-max normalized importance degrees in each test document. Figure 6 shows the standard deviations sorted by ascending order.

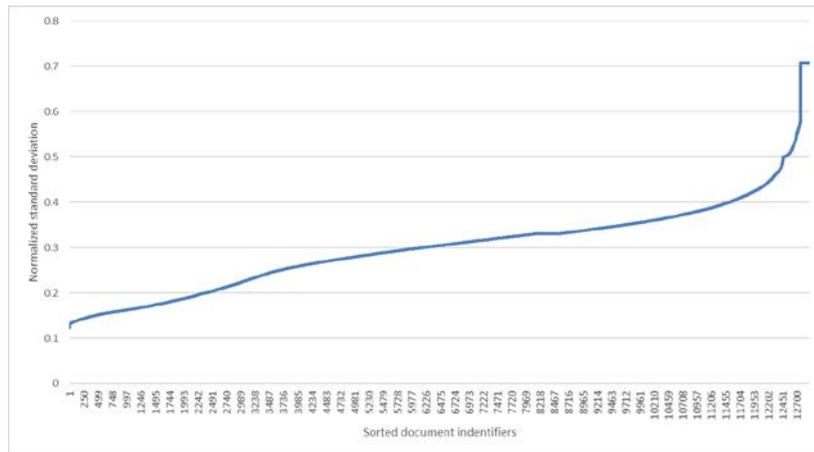

**Figure 6.** Standard deviations of importance degrees

As shown in Figure 6, the number of documents where the standard deviation is over 0.2 was 81.6%. Although the statistic is not computed from a gold standard dataset (i.e., dataset manually annotated with importance scores of each sentence), we can indirectly find that sentences in a document differently support polarity of the whole document.

When the proposed model returned incorrect polarity values, we checked the score differences between predicted polarity values and correct polarity values. Figure 7 shows the number of incorrectly predicted documents according to the score differences.

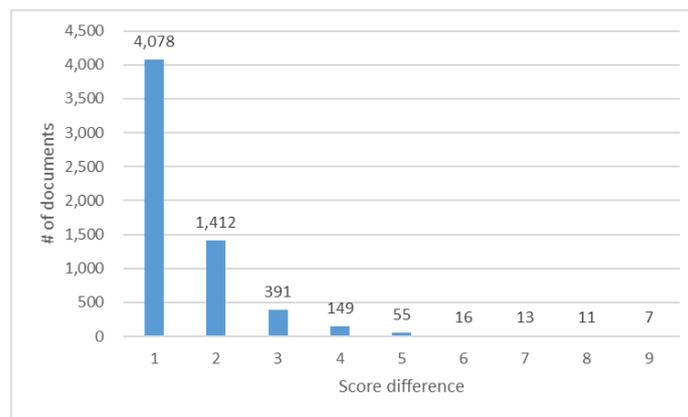

**Figure 7.** Score differences in wrong predictions

As shown in Figure 7, the number of documents where the score difference is just one point was 66.5% of all wrong predictions (4078 ones among 6132 documents). Then, 89.5% of wrong predictions (5490 ones among 6132 documents) showed the score differences within two points. In 10-scale polarity prediction, we think that these small score differences often occur even in human prediction. Therefore, various inter-coder agreement rates for IMDB dataset should be reported to measure reliability for qualitative categorical items.

## 6. Conclusion

We propose an effective neural network model for document-level sentiment classification. The proposed model automatically determines the importance degrees of sentences in documents using gate functions learned from mass training data. Then, it classifies an input document into predefined sentiment classes by differently considering the importance degrees of each sentence. In the experiments with the four different datasets, the proposed model showed better performances than previous state-of-the-art models. From the experimental results, we found that the importance of sentences should be considered in a document-level sentiment classification task. Our future work will focus on a more effective neural network architecture for reflecting sentence importance. In addition, we will focus on a light document encoder for replacing large-scale language models.

**Author contributions:** Conceptualization, Harksoo Kim; methodology, Harksoo Kim; software, Gihyeon Choi and Shinhyeok Oh; validation, Gihyeon Choi and Shinhyeok Oh; formal analysis, Harksoo Kim; investigation, Harksoo Kim; resources, Gihyeon Choi and Shinhyeok Oh; data curation, Gihyeon Choi and Shinhyeok Oh; writing—original draft preparation, Gihyeon Choi; writing—review and editing, Harksoo Kim; visualization, Harksoo Kim; supervision, Harksoo Kim; project administration, Harksoo Kim; funding acquisition, Harksoo Kim.

**Funding:** This work was supported by Institute of Information & communications Technology Planning & Evaluation(IITP) grant funded by the Korea government(MSIT) (No.2020-0-00368, A Neural-Symbolic Model for Knowledge Acquisition and Inference Techniques).

**Acknowledgments:** We especially thank the members of the NLP laboratory at Konkuk University for their technical support.

**Conflicts of Interest:** The authors declare no conflict of interest.